\documentclass{article}
\usepackage{ijcai19}

\usepackage{times}
\usepackage{soul}
\usepackage{url}
\usepackage[hidelinks]{hyperref}
\usepackage[utf8]{inputenc}
\usepackage[small]{caption}
\usepackage{graphicx}
\usepackage{amsmath}
\usepackage{booktabs}
\usepackage{amssymb}
\usepackage{mathrsfs}
\usepackage{wasysym}
\usepackage{color}
\usepackage{xspace}
\usepackage{txfonts}
\usepackage{enumerate}
\usepackage{tikz}
\usetikzlibrary{automata,arrows,positioning}
\usepackage{nicefrac}
\usepackage[algoruled,linesnumbered,noend]{algorithm2e}
\usepackage{multirow}
\usepackage{diagbox}
\usepackage{hyperref}

\newcommand{\FPath}{{S^+}}
\newcommand{\CGS}{{CGS}\xspace}

\newcommand{\last}{{\tt lst}}

\newcommand{\G}{{\cal G}\xspace}
\newcommand{\ir}{{\tt ir}\xspace}
\newcommand{\Ir}{{\tt Ir}\xspace}

\newcommand{\StrT}{\textsc{T$_{\tt str}$}\xspace}
\newcommand{\iR}{{\tt iR}\xspace}
\newcommand{\IR}{{\tt IR}\xspace}
\newcommand{\play}{{\tt play}\xspace}

\newcommand{\STCGS}{{ACGS}\xspace}

\newcommand{\M}{{\cal M}\xspace}
\newcommand{\opUA}[1]{[[{#1}]]}
\newcommand{\opA}[1]{\langle\langle{#1}\rangle\rangle}

\newcommand{\PP}{\mathcal{A}}
\newcommand{\PG}{\mathcal{P}}

\newcommand{\outcomes}{{\cal O}}

\newcommand{\nA}{\overline{A}}

\newcommand{\inff}{\textsf{inf}}
\newcommand{\opX}{{\bf X}}
\newcommand{\opK}{{\bf K}}
\newcommand{\opD}{{\bf D}}
\newcommand{\opE}{{\bf E}}
\newcommand{\opF}{{\bf F}}

\newcommand{\opC}{{\bf C}}

\newcommand{\opU}{{\bf U}}
\newcommand{\opR}{{\bf R}}
\newcommand{\opG}{{\bf G}}
\newcommand{\dom}{{\tt dom}}
\newcommand{\qed}{$\Box$}

\newtheorem{proposition}{Proposition}
\newtheorem{theorem}{Theorem}

\newtheorem{lemma}{Lemma}

\def\squareforqed{\hbox{\rlap{$\sqcap$}$\sqcup$}}
\def\qed{\ifmmode\squareforqed\else{\unskip\nobreak\hfil
\penalty50\hskip1em\null\nobreak\hfil\squareforqed
\parfillskip=0pt\finalhyphendemerits=0\endgraf}\fi}

\begin{document}

\title{Making Agents' Abilities Explicit}

\author{
Yedi Zhang$^1$
\and
Fu Song$^1$\And
Taolue Chen$^2$
\affiliations
$^1$ShanghaiTech University, Shanghai, China\\
$^2$Birkbeck, University of London, UK}

\maketitle


\begin{abstract}
  Alternating-time temporal logics (ATL/ATL$^\ast$) represent a family of modal logics for reasoning about agents' strategic abilities in multiagent systems (MAS). The interpretations of ATL/ATL$^\ast$ over the semantic model Concurrent Game Structures (\CGS) usually vary depending on the agents' abilities, for instance, perfect vs. imperfect information, perfect vs. imperfect recall, resulting in a variety of variants which have been studied extensively in literature. However, they are defined at the semantic level, which may limit modeling flexibilities and may give counter-intuitive interpretations.  
  To mitigate these issues, in this work, we propose to extend 
  \CGS with agents' abilities 
  and study the new semantics of ATL/ATL$^\ast$ under this model. We give PSACE/2EXPTIME model-checking algorithms for ATL/ATL$^\ast$ and implement them as a prototype tool. Experiment results show the practical feasibility of the approach.

\end{abstract}


\section{Introduction}
\label{sec:intr}

\emph{Multiagent systems} (MAS) consisting of multiple autonomous agents are a wide adopted paradigm of intelligent systems. 
Game-based models and associated logics, as the foundation of MAS, have received tremendous attention in recent years. The seminar work~\cite{AHK02} proposed \emph{concurrent game structures} (CGS) as the model of MAS and alternating-time temporal logics (typically ATL and ATL$^\ast$) as specification languages for expressing temporal goals. In a nutshell, a CGS consists of multiple players which are used to represent autonomous agents, components and the environment. The model describes how the MAS evolves according to the collective behavior of agents. ATL/ATL$^\ast$, an extension of the Computational Tree Logics (CTL/CTL$^\ast$), features coalition modality $\opA{A}$. The formula $\opA{A}\varphi$ expresses the property that the coalition (i.e. the agent group) $A$ has a collective strategy to achieve a certain goal specified by $\varphi$.



A series of extensions of ATL-like logics have been studied which take different agents' abilities into account. These abilities typically include whether the agents can identify the current state of the system completely or only partially (perfect vs.\ imperfect information), and whether the agents can memorize the whole history of observations or simply part of them (perfect vs.\ imperfect recall). Different abilities usually induce distinct  semantics of logics, which are indeed necessary because of the versatility of problem domains. These semantic variants and their model-checking problems comprise subjects of active research for almost two decades, to cite a few~\cite{Sch04,JH04,AGJ07,DT11,BJ11,LM15}.

While the agents' abilities play a prominent role~\cite{BJ14}, the semantics of ATL-like logics only refers to them \emph{implicitly}. In other words, the logic per se does not specify what ability an agent has; instead one can infer the ability an agent requires by examining the goal specified in the logic.
This approach, being elegant and valuable to understand the relationship between different abilities, suffers from a few shortcomings: (1) From the modelling perspective, it is common in practice that agents in an MAS vary in their abilities (for instance, agents modeling sensors may not identify the complete state of system so can only use strategies with imperfect information). When building up a model, these abilities ought to be encoded explicitly. Such modeling flexibility is not supported by existing formalisms.
(2) From the semantic perspective, ATL-like logics may exhibit some counter-intuitive semantics.
$\opA{A}\varphi$ is interpreted as the coalition $A$ has a collective strategy to achieve the goal $\varphi$ ``no matter
what the other agents do" rather than ``no matter which strategies the other agents choose",
hence, neglects the (multi-player) game nature in the evolution of MAS.
For instance, when it comes to the imperfect information/recall setting, only agents in the formula $\opA{A}\varphi$ are assumed to use imperfect information/recall strategies, while the agents \emph{not} in $A$ may still use perfect information and perfect recall strategies. Even worse, if the coalition modalities are nested, the same agent may have different abilities to fulfill the objectives specified in different sub-formulae, which results in inconsistency.
This phenomenon has also been mentioned in~\cite{MMPV14,CermakLMM14} which proposed a strategic logic does make explicit references to strategies of all agents including those not in $A$.
However, all agents in strategic logic should have same abilities.

To summarize, it occurs to us that the current approach in which the temporal formulae are with implicit agents' abilities at the semantics level impedes necessary modeling flexibility and often yields unpleasant (even weird) semantics. Instead, we argue that coupling agents' abilities at the syntactic level of system models would deliver a potentially better approach to overcome the aforementioned limitations. Bearing the rationale in mind, we propose a new MAS model, named \emph{Agents' Abilities Augmented Concurrent Game Structures} (\STCGS), which encompasses agents' abilities explicitly.

We investigate ATL/ATL$^\ast$ over \STCGS. We show that in general the new semantics of ATL/ATL$^\ast$ over \STCGS is incomparable with others even if the underlying CGS models are the same.
We also study the model-checking problem of ATL/ATL$^\ast$ over \STCGS. We show that this problem is generally undecidable, as the problem of ATL over \CGS under the imperfect information and perfect recall setting is already undecidable~\cite{DT11}. However, we manage to show that the model-checking problem for ATL$^\ast$ (resp. ATL) on \STCGS is 2EXPTIME-complete (resp. in PSPACE) when the imperfect information and perfect recall strategies are disallowed. We implement our algorithms in a prototype tool and conduct experiments on some standard applications from the literature. The results confirm the feasibility of our approach.

The source code of our tool is available at~\cite{ACGStool19} which also includes
some further experiments and comparison of ATL/ATL$^\ast$ semantics
between CGS and \STCGS.

\section{Concurrent Game Structures}
Given an infinite word $\rho=s_0s_1\ldots$, we denote by $\rho_j$ the symbol $s_j$,
$\rho_{[0..j]}$ the prefix $s_0\ldots s_j$, and $\rho_{[j..\infty]}$ the suffix $s_js_{j+1}\ldots$.
Given a finite word $\rho=s_0s_1\cdots s_m$, we denote by $\rho_j$ the symbol $s_j$ for $0\leq j\leq m$, and
$\last(\rho)$ the symbol $s_m$.

Let $AP$ denote a finite set of atomic propositions. A \emph{concurrent game structure} (\CGS) is a tuple
\begin{center}
    $\G\triangleq(S, S_0, Ag, (Ac_i)_{i\in Ag}, (\sim_i)_{i\in Ag}, (P_i)_{i\in Ag}, \Delta, \lambda )$,
\end{center}
where $S$ is a finite set of \emph{states};
$S_0\subseteq S$ is a set of \emph{initial states};
$Ag=\{1,...,n\}$ is a finite set of \emph{agents};
$Ac_i$ is a finite set of \emph{local actions} of agent $i$;
$\sim_i\subseteq S\times S$ is an \emph{epistemic accessibility relation} (an equivalence relation);
$P_i: S \rightarrow 2^{Ac_i}$ is a \emph{protocol function} such that $P_i(s)=P_i(s')$ for every $s\sim_i s'$;
$\Delta: S\times Ac\rightarrow S$ is a transition function with $Ac=\prod_{i\in Ag} Ac_i$ being a set of joint actions; and $\lambda:S\rightarrow 2^{AP}$ is a \emph{labeling function} which assigns each state with a set of atomic propositions.
Given a joint action $\vec{a}=\langle a_1,...,a_n\rangle\in Ac$, we use $\vec{a}(i)$ to denote the local action of agent $i$ in $\vec{a}$.
%
%

A \emph{path} is an infinite sequence of states $\rho=s_0s_1\ldots$ such that for every $j\geq 0$, $s_{j+1}=\Delta(s_j, \vec{a}_j)$ for some $\vec{a}_j\in \prod_{i\in Ag}P_i(s_j)$.
Two sequences $\rho=s_0\ldots s_m\in S^+$ and $\rho'=s_0'\ldots s_m'\in S^+$ are \emph{indistinguishable} for agent $i$, denoted by $\rho\sim_i \rho'$, if for every $j:0\leq j\leq m$, $s_j\sim _i s_j'$.

 \smallskip
 \noindent
 \textbf{Strategies.} 
Typical agents' abilities are captured by the following types of strategies~\cite{Sch04}. For $i\in Ag$,
 \begin{description}
   \item[\Ir-strategy] $\theta_i: S \rightarrow Ac$ with $\forall s\in S$, $\theta_i(s)\in P_i(s)$; 
   \item[\IR-strategy] $\theta_i:\FPath\rightarrow Ac$ with $\forall\rho\in\FPath$, $\theta_i(\rho)\in P_i(\last(\rho))$; 
   \item[\ir-strategy] $\theta_i:S\rightarrow Ac$, the same as the \Ir-strategy but with the additional constraint $s\sim_i s'\Rightarrow\theta_i(s)=\theta_i(s')$; 
   \item[\iR-strategy] $\theta_i:\FPath\rightarrow Ac$, the same as the \IR-strategy but with the additional constraint $\rho\sim_i \rho'\Rightarrow\theta_i(\rho)=\theta_i(\rho')$. 
 \end{description}
%
%
%
Intuitively, ${\tt i}$ (resp. ${\tt I}$) signals that agents can only observe partial information characterized via epistemic accessibility relations (resp. complete information with all epistemic accessibility relations being the identity relation), while ${\tt r}$ (resp. ${\tt R}$) signals that agents can make decision based on the current observation (resp. the whole history of observations). We will, by slightly abusing notation, extend \Ir-strategies and \ir-strategies $\theta_i$ to the domain $\FPath$ such that for all $\rho\in\FPath$, $\theta_i(\rho)=\theta_i(\last(\rho))$. We denote by $\Theta_i^\sigma$ for $\sigma\in\{\Ir,\IR,\ir,\iR\}$ the set of $\sigma$-strategies for agent $i$.

\smallskip
\noindent
{\bf Outcomes.} A \emph{collective $\sigma$-strategy} of a set of agents $A$ is a function $\upsilon_A^\sigma$ assigning each agent $i\in A$ with a $\sigma$-strategy $\upsilon_A^\sigma(i)\in \Theta_i^\sigma$. For $i\in A$ and $\rho\in\FPath$, we denote the local action $\upsilon_A^\sigma(i)(\rho)$ of agent $i$ by  $\upsilon_A^\sigma(i,\rho)$, and the set $Ag\setminus A$ by $\nA$.

Given a state $s$, two collective $\sigma/\sigma'$-strategies $\upsilon_A^\sigma$ and $\upsilon_{\nA}^{\sigma'}$ yield a path $\rho$, denoted by $\play(s,\upsilon_A^\sigma,\upsilon_{\nA}^{\sigma'})$, where $\rho_0=s$ and for every $j\geq 0$, $\rho_{j+1}=\Delta(\rho_j,\vec{a}_j)$, $\vec{a}_j(i)=\upsilon_A^\sigma(i,\rho_{[0..j]})$ for $i\in A$ and $\vec{a}_j(i)=\upsilon_{\nA}^{\sigma'}(i,\rho_{[0..j]})$ for $i\in \nA$.

For every state $s\in S$ and collective $\sigma$-strategy $\upsilon_A^\sigma$ of $A$, the \emph{outcome} function is defined as follows:
  \begin{center}
   $\outcomes_\G^\sigma(s,\upsilon_A^\sigma)\triangleq \{\play(s,\upsilon_A^\sigma,\upsilon_{\nA}^{\IR})\mid \forall i\in\nA,\upsilon_{\nA}^{\IR}(i)\in\Theta_i^{\IR} \}$,
  \end{center}
  i.e., the set of all possible plays that may occur when each agent $i\in A$ enforces its $\sigma$-strategy $\upsilon_A^\sigma(i)$ from the state $s$.
The subscript $\G$ is dropped in $\outcomes_\G^\sigma$ when it is clear from the context.

\section{Alternating-Time Temporal Logics}
\label{sec:logic}

The alternating-time temporal logic ATL/ATL$^\ast$ is an extension of the branching-time logic CTL/CTL$^\ast$ by replacing the existential path quantifiers ${\bf E}$ with coalition modalities $\opA{A}$~\cite{AHK02}. Intuitively, the formula $\opA{A}\phi$ expresses that the set of agents $A$ has a collective strategy to achieve the goal $\phi$ no matter which strategies the agents in $\nA$ choose.
Formally,  ATL$^\ast$ is defined by the following grammar:
  \begin{align*}
    \varphi::= & \ q \ \mid \neg\varphi \ \mid \varphi \wedge \varphi \ \mid \ \opA{A}\phi \\
    \phi::=  & \ \varphi \ \mid\   \neg \phi\ \mid\ \phi \wedge \phi \ \mid\  \opX\ \phi \  \mid\ \phi\ \opU\ \phi
  \end{align*}
where $\varphi$ (resp. $\phi$) denotes state (resp. path) formulae, $q\in AP$ and $A\subseteq  Ag$.

The derived operators are defined as usual: $\phi_1\rightarrow \phi_2 \triangleq\phi_2\vee\neg\phi_1$, $\opF \ \phi \triangleq {\tt true} \ \opU \ \phi$, $\opG \ \phi \triangleq \neg \opF \ \neg \phi$, $\phi_1\ \opR\ \phi_2\triangleq\opG \phi_2\vee \phi_2\opU (\phi_1\wedge \phi_2)$, and $\opUA{A}\phi \triangleq \neg\opA{A}\neg\phi$.  An LTL formula is an ATL$^\ast$ path formula by restricting $\varphi$ to atomic propositions.

The semantics of ATL$^\ast$ is traditionally defined over \CGS. When agents's abilities are considered, it is often parameterized with a strategy type $\sigma\in\StrT$, denoted by ATL$^\ast_\sigma$ \cite{BJ14}.
 Formally, let $\G$ be a \CGS and $s$ be a state of $\G$, the semantics of ATL$^\ast_\sigma$ (i.e. the satisfaction relation) is defined inductively as follows:
(where $\varrho$ is a state $s$ or path $\rho$)
 \begin{itemize}
   \item $\G, s \models_\sigma q$ iff $q\in\lambda(s)$;
   \item $\G, s \models_\sigma \opA{A}\phi$ iff there exists a collective $\sigma$-strategy $\upsilon_A^\sigma$ of agents $A$ such that $\forall \rho\in \outcomes^\sigma(s,\upsilon_A^\sigma)$, $\G, \rho \models_\sigma \phi$;
   \item $\G, \rho \models_\sigma \varphi$ iff $\G, \rho_0 \models_\sigma \varphi$;
   \item $\G,\rho\models_\sigma \opX\phi$ iff $\G, \rho_{[1..\infty]} \models_\sigma \phi$;
   \item $\G,\rho\models_\sigma \phi_1\opU\phi_2$ iff $\exists k\geq 0$ such that $\G, \rho_{[k..\infty]} \models \phi_2$ and $\forall j: 0\leq j < k$, $\G, \rho_{[j..\infty]}\models_\sigma \phi_1$;
   \item $\G, \varrho \models_\sigma \phi_1\wedge \phi_2$ iff  $\G, \varrho \models_\sigma \phi_1$ and $\G, \varrho \models_\sigma \phi_2$;
   \item $\G, \varrho \models_\sigma \neg \phi$ iff $\G, \varrho \not\models_\sigma \phi$.
 \end{itemize}

\medskip
 \noindent
 {\bf Vanilla ATL}. ATL is a sublogic of ATL$^\ast$ where each occurrence of the coalition modality $\opA{A}$ is immediately followed by a temporal operator. Formally, ATL is defined by the following grammar:\\
    $\varphi::= q  \mid \neg \varphi   \mid \varphi \wedge \varphi   \mid \opA{A}\opX \ \varphi  \mid    \opA{A}[\varphi  \opR  \varphi]  \mid   \opA{A}[\varphi \ \opU  \ \varphi]$
 where $q\in A$ and $A\subseteq Ag$.

 Remark that the operator $\opR$ cannot be defined using other operators in ATL with imperfect information~\cite{LMO08}, so is included for completeness.

Given an ATL$^\ast$ formula $\varphi$, a CGS $\G$ and a strategy type $\sigma\in\StrT$,
the \emph{model-checking problem} is to determine whether $\G, s \models_\sigma \varphi$ or not, for each initial state $s$ of the CGS $\G$.

\medskip
\noindent
{\bf Some Semantic Issues.}
We observe that the semantics of ATL/ATL$^\ast$ refers to the agents' abilities in an implicit manner. 
For the formula $\opA{A}\varphi$, the specified $\sigma$-strategies only apply to 
agents in $A$ 
while the agents in $\nA$ could still choose beyond $\sigma$-strategies (e.g. $\IR$-strategies). In other words,
$A$ has a collective $\sigma$-strategy to achieve $\varphi$ no matter what the other agents do.
When $\sigma$ is $\IR$ as in the original work by~\cite{AHK02},  this interpretation
of $\opA{A}\varphi$ is plausible, as ``no matter what the other agents do" is effectively the same as
``no matter which strategies the other agents choose".
However, when $\sigma$ is set to be  
more restricted than $\IR$,
agents not in $A$ are still allowed to use $\IR$-strategies.

As mentioned in the introduction, this results in a few shortcomings.
From a modeling perspective,
arguably agents' abilities should be decided by the practical scenario. Namely, they should be fixed when the model is built, and all agents stick to their respective abilities independent of logic formulae.
More concretely, from the semantic perspective, the existing semantics does not take into account the abilities of agents who are not in $A$, and neglects the (multi-player) game nature in the evolution of MAS. As a result, it
may exhibit some counter-intuitive semantics.
For instance, consider two formulae $\opA{A}\phi$ and $\opA{A'}\phi'$ such that agent $i \in A\setminus A'$,
$i$ may have different abilities to achieve $\phi$ and $\phi'$. Let us consider an autonomous road vehicle scenario to see why this is not ideal. There are several autonomous cars which can only observe partial information and have bounded memory. An MAS model $\G$ consists of agents in $A$ modeling autonomous cars, and an additional environment agent $e$. We can reasonably assume that all the car agents use $\ir$-strategies, while $e$ uses $\IR$-strategies.
The property $\opA{A'}\phi$ expresses that autonomous cars $A'\subset A$ can cooperatively achieve the goal $\phi$ no matter what strategies the other cars and the environment choose. Verifying that $\G$ satisfies $\opA{A'}\phi$ under the existing semantics would allow car agents $A\setminus A'$ to use $\IR$-strategies.
If $\G$ satisfies $\opA{A'}\phi$, then the result is conclusive, i.e., $\opA{A'}\phi$ holds for the system.
However, if $\G$ invalidates $\opA{A'}\phi$, we \emph{cannot} deduce that $\opA{A'}\phi$ fails because we overestimate the abilities of agents in $A\setminus A'$ when evaluating $\opA{A'}\phi$. 
In other words, for the formula $\opA{A}\varphi$ under $\models_\sigma$ where  $\sigma\neq \IR$, it seems to be inappropriate to render the agents not in $A$ extra powers of $\IR$ to potentially defeat agents from $A$ when their abilities are actually much weaker.

 \section{Agents' Abilities Augmented Concurrent Game Structures}
 \label{sec:STCGS}

In this section, we introduce \emph{agents' abilities augmented concurrent game structures} (\STCGS in short), which explicitly equip each agent with a strategy type from $\StrT$. As such, agents have fixed abilities throughout their lives for a given CGS. Formally, an \STCGS is a pair $\M \triangleq (\G, \pi)$, where $\G$ is a CGS and $\pi:Ag\rightarrow \StrT$ is a function that assigns a strategy type $\pi(i)$ to the agent $i$. We assume that, for each agent $i\in Ag$ with  $\pi(i)\in \{\IR,\Ir\}$, $\sim_i$ is an identity relation, as agents with perfect information should be able to distinguish two distinct states.
Paths of $\M$ are defined the same as for the CGS $\G$, but strategies and outcomes of $\M$ have to be redefined as follows.

\smallskip
\noindent
{\bf Strategies and Outcomes.}
Let $A$ be a set of agents.  A \emph{collective strategy} of $A$ in $\M$ is a function $\xi_A$ that assigns each agent $i\in A$ with a $\pi(i)$-strategy $\xi_A(i)\in \Theta_i^{\pi(i)}$.

 Given a state $s\in S$, for every collective strategy $\xi_A$ of the set $A$ of agents, the outcome $\outcomes_\M(s,\xi_A)$ of $\M$ is the set of all possible paths that  may occur when each agent $i\in A$ enforces its $\pi(i)$-strategy $\xi_A(i)$ from state $s$, and other agents $i\in\nA$ can only choose $\pi(i)$-strategies instead of general $\IR$-strategies. Formally, $\outcomes_\M(s,\xi_A)$ is defined as
 \[
    \outcomes_\M(s,\xi_A) \triangleq \{\play(s',\xi_A,\xi_{\nA})\mid \forall i\in\nA,\xi_{\nA}(i)\in {\Theta_{i}^{\pi(i)}}\}
 \]
We will omit the subscript $\M$ from $\outcomes_\M(s,\xi_A)$ when it is clear from context.

 \smallskip
 \noindent
 {\bf Semantics of ATL/ATL$^\ast$.} The difference of outcomes between \STCGS and \CGS induces a distinct semantics of ATL$^\ast$ on \STCGS than \CGS. Let $\M$ be an \STCGS and $s$ be a state in $\M$, the semantics of ATL$^\ast$ on $\M$ is defined similar to the one on \CGS, except that the semantics of the state formulae of the form $\opA{A}\phi$ is defined as follows:

 \begin{quote}
   $\M, s \models \opA{A}\phi$ iff there exists a collective strategy $\xi_A$ of $A$ such that $\M, \rho \models \phi$ for all $\rho\in \outcomes(s,\xi_A)$.
 \end{quote}

Remark that this semantics takes into account whether the agents from $\nA$ have perfect or imperfect information/recall.

Given an \STCGS $\M$ and an ATL/ATL$^\ast$ formula $\varphi$,
the \emph{model-checking problem} is to determine whether $\M, s \models \varphi$ holds, for every initial state $s$ of $G$.
Given a state formula $\varphi$, let $\llbracket \varphi\rrbracket_\M$ denote the set of all the states of $\M$ that satisfies $\varphi$.

 The semantics of ATL/ATL$^\ast$ defined on \STCGS is different from the one defined on \CGS, hence they are incomparable.

 \begin{proposition}\label{prop:stcgs2cgs}
 There are an \STCGS $\M=(\G,\pi)$, an ATL/ATL$^\ast$ formula $\opA{A}\phi$, and  a type $\sigma\in\StrT$ such that $\pi(i)=\sigma$ for all $i\in A$ and $\M,s\models \opA{A}\phi$ holds, but $\G,s\not\models_\sigma \opA{A}\phi$.
 \end{proposition}
 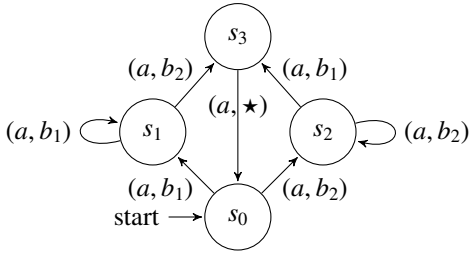
\begin{figure}[t]
  \centering
  \begin{tikzpicture}[->,>=stealth',shorten >=0.8pt,node distance=1.6cm]
    \node[state,initial] (s0) {$s_0$};
    \node[state] (s1) [above left of=s0] {$s_1$};
    \node[state] (s2) [above right of=s0] {$s_2$};
    \node[state] (s3) [above of=s0,node distance=2.4cm] {$s_3$};

    \path[->]
      (s0) edge node [left, xshift=0.1cm,yshift=-0.2cm] {$(a,b_1)$} (s1)
      (s0) edge node [right, xshift=-0.1cm,yshift=-0.2cm] {$(a,b_2)$} (s2)
      (s1) edge[loop left] node {$(a,b_1)$} ()
      (s1) edge node [left,  xshift=0.1cm,yshift=0.2cm]{$(a,b_2)$} (s3)
      (s3) edge node [above] {$(a,\star)$} (s0)
      (s2) edge[loop right] node {$(a,b_2)$} ()
      (s2) edge node [right, xshift=-0.1cm,yshift=0.2cm]{$(a,b_1)$} (s3)       ;
  \end{tikzpicture}
  \caption{An illustrating example, where $\star\in\{b_1,b_2\}$.}
  \label{fig:exm1}
 \end{figure}
\emph{Proof.} Let us consider the \CGS shown in Figure~\ref{fig:exm1}. There are two agents $\{1,2\}$, four states $\{s_0,s_1,s_2,s_3\}$ ($s_0$ is the initial state), $\lambda(s_0)=\lambda(s_1)=\lambda(s_2)=\{q\}$ and $\lambda(s_3)=\emptyset$, $\sim_1$ is the identity relation, $s\sim_2s'$ for every $s,s'\in\{s_0,s_1,s_2\}$ and $s_3\sim_2s_3$. Consider the function
 $\pi$ such that $\pi(1)=\IR$ and $\pi(2)=\ir$, it is easy to see that $\M,s_0\models \opA{\{1\}}\opG q$, while $\G,s_0\not\models_\IR \opA{\{1\}}\opG q$.
 \qed
%

%

\section{Model-Checking Algorithms}
 \label{sec:mc}

It has been shown that the Turing Halting problem can be  reduced to the model-checking problem of \CGS against the ATL formula $\varphi=\opA{\{1,2\}}\opG  \ ok$ under the $\iR$ setting~\cite{DT11}, where
$ok$ is an atomic proposition. By adapting the proof, we get that: 

 \begin{theorem} \label{thm:und}
   The model-checking problem for \STCGS $\M=(\G,\pi)$ against the ATL/ATL$^\ast$ formula $\opA{\{1,2\}}\opG  \ ok$ is undecidable, where $\{1,2,3\}\subseteq Ag$,
   $\pi(1)=\pi(2)=\iR$ and $\pi(i)=\IR$ for $i\in Ag\setminus\{1,2\}$.
 \end{theorem}


By Theorem~\ref{thm:und}, we focus on the model-checking problem of \STCGS by restricting the function $\pi$ to $Ag\rightarrow \StrT\setminus\{\iR\}$.
In general, we propose model-checking algorithms which iteratively compute the set of states satisfying state formulae from the innermost subformulae. The main challenge is to compute $\llbracket \opA{A}\phi\rrbracket_\M$. To this end, we first show how to compute $\llbracket \opA{A}\phi\rrbracket_\M$ for a simple formula $\opA{A}\phi$,
and then present the more general algorithm. An ATL/ATL$^\ast$ formula $\opA{A}\phi$ is called \emph{simple} if $\phi$ is an LTL formula.

Let us fix an \STCGS $\M= (\G, \pi)$ with $\G=(S,S_0,Ag,(Ac_i)_{i\in Ag},(\sim_i)_{i\in Ag},(P_i)_{i\in Ag},\Delta,\lambda)$
and a simple formula $\opA{A}\phi$.
Given a set of agents $A'$ and a strategy type $\sigma\in \StrT$,
we denote by $A'_{\sigma}$ the set $\{i\in A'\mid \pi(i)=\sigma\}$.

\subsection{Model-Checking Simple ATL Formulae}
For a simple ATL formula $\opA{A}\phi$, it is easy to see that whether agents in $A$ have perfect call or not does not matter if these agents have perfect information abilities.

 \begin{proposition}\label{prop:Isemid}
   Given an \STCGS $(\G,\pi)$ with $\pi:Ag\rightarrow \StrT\setminus \{\iR\}$, and a simple ATL formula $\opA{A}\varphi$,
   let $\pi'$ be a function such that for every $i\in Ag$, $\pi'(i)=\Ir$ if $\pi(i)=\IR$ and $i\in A$, otherwise $\pi'(i)=\pi(i)$. For every state $s$ in $\M$,
   \begin{center}
     $(\G,\pi), s \models \opA{A}\varphi$ iff $(\G,\pi'), s \models \opA{A}\varphi$.
   \end{center}
 \end{proposition}

By Proposition~\ref{prop:Isemid}, all the agents in $A$ with $\IR$-strategies can be seen as agents with $\Ir$-strategies.
All the agents in $Ag$ with $\Ir$-strategies can be seen as agents with $\ir$-strategies (i.e., all the epistemic
accessibility relations of them are the identity relation).
Therefore, we can assume that $\pi(i)=\ir$ for all $i\in A$,
and $\pi(i)\in\{\ir,\IR\}$ for all $i\in \nA$.
For two collective strategies $\xi_A$ and $\xi_{\nA_{\ir}}$,
let $\M(\xi_A,\xi_{\nA_{\ir}})=(\G',\pi)$ be the \STCGS obtained from $(\G,\pi)$ by enforcing strategies $\xi_A$ and $\xi_{\nA_{\ir}}$,
namely, by removing transitions whose actions of agents in $A\cup \nA_{\ir}$ do not conform to $\xi_A$ and $\xi_{\nA_{\ir}}$. We have that
\begin{center}
$\llbracket \opA{A}\phi\rrbracket_\M\equiv \bigcup_{\xi_A} \bigcap_{\xi_{\nA_{\ir}}}\llbracket
\opA{\emptyset}\phi\rrbracket_{\M(\xi_A,\xi_{\nA_{\ir}})}$.
\end{center}

Computing $\llbracket \opA{\emptyset}\phi\rrbracket_{\M(\xi_A,\xi_{\nA_{\ir}})}$ amounts to CTL model-checking,
which can be done in polynomial time (and thus in polynomial space) in the size of $\M(\xi_A,\xi_{\nA_{\ir}})$ and $\opA{\emptyset}\phi$~\cite{CES83}.
Since the number of strategies $\xi_A$ and $\xi_{\nA_{\ir}}$  is finite, we get that:

%
%
\begin{lemma}\label{lem:guessIr}
For the simple ATL formula $\opA{A}\phi$, $\llbracket \opA{A}\phi\rrbracket_\M$ can be computed in PSPACE.
\end{lemma}


\subsection{Model-Checking Simple ATL$^\ast$ Formulae}
We compute $\llbracket \opA{A}\phi\rrbracket_\M$ by a reduction to the problem of computing the winning region of a turn-based two-player parity game. We first introduce some basic concepts which will be used in our reduction.

A \emph{deterministic parity automaton} (DPA) is a tuple $\PP=(P,\Sigma,\delta, p_0, R)$, where $P$ is a finite set of states, $\Sigma$ is the input alphabet, $\delta: P\times\Sigma \rightarrow P$ is a transition function, $p_0\in P$ is the initial state and $R:P \rightarrow\{0,...,k\}$ is a \emph{rank} function. A \emph{run} $\rho$ of $\PP$ over an $\omega$-word $\alpha_0\alpha_1...\in \Sigma^\omega$ is an infinite sequence of states $\rho=p_0p_1...$ such that for every $i\geq 0$, $p_{i+1}=\delta(p_i,\alpha_i)$. Let $\inff(\rho)$ be the set of states visited infinitely often in $\rho$.
 An infinite word is \emph{recognized} by $\PP$ if $\PP$ has a run $\rho$ over this word
 such that $\min_{p\in\inff(\rho)} R(p)$ is \emph{even}. For the LTL formula $\phi$, one can construct a DPA $\PP_\phi=(P,2^{AP},\delta, p_0, R)$ with $2^{2^{O(|\phi|)}}$ states and rank $k=2^{O(|\phi|)}$ such that $\PP_\phi$ recognizes all the $\omega$-words satisfying $\phi$ \cite{Pit06}.

A \emph{(turned-based, two-player) parity game} $\PG$ is a tuple $(V=V_0\uplus V_1,E, \Xi)$, where $V_i$ for $i\in\{0,1\}$ is a finite set of vertices controlled by Player-$i$,
$E\subseteq V\times V$ is a finite set of edges, and $\Xi:V\rightarrow\{0,...,k\}$ is a \emph{rank} function.
A \emph{play} $\rho$ starting from $v_0$ is an infinite sequence of vertices $v_0v_1...$
such that for every $i\geq 0$, $(v_i,v_{i+1})\in E$.
A \emph{strategy} of Player-$i$ is a function $\theta:V^*V_i\rightarrow V$
such that for every $\rho\in V^*$ and $v\in V_i$, $(v,\theta(\rho\cdot v))\in E$.
Given a strategy $\theta_0$ for Player-0 and a strategy $\theta_1$ for Player-1,
let $G_{\theta_0,\theta_1}$ be the play where Player-0 and Player-1 enforce their strategies $\theta_0$ and $\theta_1$.
$\theta_0$ is a winning strategy for Player-0 
if $\min_{s\in\inff(G_{\theta_0,\theta_1})} \Xi(s)$ is \emph{even} for every strategy $\theta_1$ of Player-1.
The winning region of Player-0, denoted by $WR_0$, is the set of vertices from which Player-0 has a winning strategy.


A \emph{partial strategy} of $A'$ is a partial function $f:A'\times S\rightarrow \bigcup_{i\in A'}Ac_i$ such that for each $i\in A'$ and $s\in S$, if $f(i,s)$ is defined, then for all $s'\in S$ with $s\sim_i s'$, $f(i,s)=f(i,s')\in P_i(s)$. We denote by $\dom(f)$ the domain of $f$. Let $F_{\ir}$ (resp. $G_{\ir}$) be the set of partial strategies of $A_{\ir}$ (resp. $\nA_{\ir}$). Let $F_{\ir}^\top$ denote the set $\{f\in F_{\ir}\mid \dom(f)=A_{\ir}\times S\}$, and $g_\bot\in G_{\ir}$ denote the partial strategy such that $\dom(g_\bot)=\emptyset$.
Given a state $s$, let $F_{\IR}^s$ be the set of functions $f:A_{\IR}\rightarrow \bigcup_{i\in A_{\IR}} Ac_i$ such that for every $i\in A_{\IR}$, $f(i)\in P_i(s)$. Let $F_{\IR}:=\bigcup_{s\in S}F_{\IR}^s$ and $\Pi_{\ir}:=\{G\subseteq G_{\ir}\mid \forall g,g'\in G.~\dom(g)=\dom(g')\}$.

We define a parity game  $\PG_\phi\triangleq(V=V_0\uplus V_1, E ,\Xi)$, where
$V_0=S\cup (S\times  P\times F_{\ir}^\top\times  \Pi_{\ir})$, $V_1=(S\times F_{\ir}^\top)\cup (S\times  P\times F_{\ir}^\top\times F_{\IR} \times \Pi_{\ir})$,
$\Xi:V\rightarrow \{0,\cdots,k\}$ is a function such that for every $s\in S$,
     \begin{itemize}
       \item  $\Xi(s)=\Xi(s,f_\top)=0$, $\forall f_\top\in F_{\ir}^\top$,
       \item  $\Xi(s,p,f_\top,G)=\Xi(s,p,f_\top,f,G)=R(p)$, $\forall p\in P$, $\forall f_\top\in F_{\ir}^\top$, $\forall f\in F_{\IR}$  and $\forall G\in \Pi_{\ir}$,
     \end{itemize}
$E$ is defined as follows:
     \begin{itemize}
       \item $(s,(s,f_\top))\in E$ for $s\in S$ and $f_\top\in F_{\ir}^\top$;
       \item $((s,f_\top), (s,p_0,f_\top,\{g_\bot\}))\in E$ for  $s\in S$ and $f_\top\in F_{\ir}^\top$;
       \item $((s,p,f_\top,G),(s,p,f_\top,f,G))\in E$ for  $(s,p,f_\top,G)\in V_0$ and $f\in F_{\IR}^s$;
%
       \item $((s,p,f_\top,f,G),(s',\delta(p,\lambda(s)),f_\top,G'))\in E$ for every $(s,p,f_\top,f,G)\in V_1$ and $s'\in S$, where $G'\subseteq \Pi_{\ir}$ is the largest set
       such that the follows hold: for every $g'\in G'$,
       \begin{enumerate}
         \item there exists $g\in G$ such that $\dom(g')=\dom(g)\cup \{(i,s'')\in \nA_{\ir}\times S\mid s\sim_i s''\}$ and for every $(i,s'')\in \dom(g)$, $g'(i,s'')=g(i,s'')$;
         \item there exists $\vec{a}\in Ac$ such that $s'=\Delta(s,\vec{a})$, and $\forall i\in Ag, ((i,s)\in \dom(f_\top)\Rightarrow \vec{a}_i=f_\top(i,s))\wedge (i\in \dom(f)\Rightarrow \vec{a}_i=f(i))\wedge((i,s)\in \dom(g')\Rightarrow \vec{a}_i=g'(i,s))$.
       \end{enumerate}
   \end{itemize}

In this reduction, $f_\top$ encodes a collective $\ir$-strategy of agents in $A_{\ir}$,
the collection of $f$'s in one play of $\PG_\phi$ encodes a collective $\IR$-strategy of agents in $A_{\IR}$,
and each $g\in G$ encodes a collective $\ir$-strategy of agents in  $\nA_{\ir}$.
The imperfect information abilities of agents are ensured by the definitions partial strategies.

To check whether $s\in \llbracket \opA{A}\phi\rrbracket_\M$,
$\PG_\phi$ starts with the vertex $s$. At first step,
Player-0 chooses a function $f_\top\in F_{\ir}^\top$ meaning that
all agents in $A_{\ir}$ choose an $\ir$-strategy.
Next, $\PG_\phi$ moves from $(s,f_\top)$ to $(s,p_0,f_\top,\{g_\bot\})$
which let the DPA $\PP_\phi$ start with $p_0$ (note that Player-1 has only one choice at this step).
At a vertex $(s,p,f_\top,G)$ controlled by Player-0, Player-0 chooses actions
for agents in $A_{\IR}$ by choosing one function $f\in F_{\IR}^s$.
Then Player-1 chooses actions for agents in $\nA$ with respect to the chosen actions of agents in $\nA_{\ir}$ tracked by $G$.
These selections of actions together with $f_\top$ and $G$ determine a joint action $\vec{a}$, based on which
$\PG_\phi$ moves to $(s',\delta(p,\lambda(s)),f_\top,G')$ such that
$s'$ is the next state of $s$ under $\vec{a}$,
$\delta(p,\lambda(s))$ is the next state of $p$ in $\PP_\phi$ which allows
to mimics the run of $\PP_\phi$ over the $\omega$-word induced by the play
of $\M$. During this step, $f$ is dropped from the vertex of $\PG_\phi$,
as $f$ corresponds to actions of agents in $A_{\IR}$ and needs not to track.
The actions of agents in $\nA_{\ir}$ are tracked by computing $G'$
from $G$. This ensures imperfect recall abilities of agents in $\nA_{\ir}$.
We then can get that:

\begin{lemma}$WR_0\cap S=\llbracket \opA{A}\phi\rrbracket_\M$.
\end{lemma}

The winning region of Player-0 in $\PG_\phi$ can be computed in polynomial time of $|V|\cdot |E|\cdot 2^k$~\cite{Jur00}.
In this reduction, each $G$ contributes at most ${O}(|S|)$ sets of $G'$. Therefore, $|V|\cdot |E|$ is exponential in $|\G|\cdot 2^{|\phi|}$.
Recall that $k=2^{O(|\phi|)}$. Consequently, we have
 \begin{lemma}
 \label{lem:sat2pg}
For the simple ATL$^\ast$ formula $\opA{A}\phi$, $\llbracket \opA{A}\phi\rrbracket_\M$ can be computed in 2EXPTIME.
 \end{lemma}


\subsection{The Overall Algorithm}
We now present the overall procedure, which computes $\llbracket \varphi\rrbracket_\M$ from the innermost subformulae. Algorithm~\ref{alg} shows the pseudo code, which takes an \STCGS $\M=(\G,\pi)$ and an ATL/ATL$^\ast$ formula $\varphi$ as inputs, and outputs $\llbracket \varphi\rrbracket_\M$. Then $\M$ satisfies $\varphi$ iff the set of initial states of $\M$ is a subset of $\llbracket \varphi\rrbracket_\M$. We also incorporate epistemic modalities $\opK_i \varphi, \opE_A\varphi, \opD_A\varphi, \opC_A\varphi$ from~\cite{HW03,CermakLMM14} into our algorithm with the following semantics:
 \begin{itemize}
   \item $\G,s\models_\sigma \opK_i\varphi ~ \mbox{iff}~\forall s'\in S,~s\sim_i s' \Longrightarrow\G, s' \models_\sigma \varphi$; 
   \item $\G,s\models_\sigma \opE_A\varphi ~ \mbox{iff}~\forall s'\in S,~s\sim_A^E s' \Longrightarrow\G, s' \models_\sigma \varphi$; 
   \item $\G,s\models_\sigma \opD_A\varphi ~ \mbox{iff}~\forall s'\in S,~s\sim_A^D s' \Longrightarrow\G, s' \models_\sigma \varphi$; 
   \item $\G,s\models_\sigma \opC_A\varphi ~ \mbox{iff}~\forall s'\in S,~s\sim_A^C s' \Longrightarrow\G, s' \models_\sigma \varphi$; 
 \end{itemize}
where $\varphi$ is a state formula, $\sim_A^E =\bigcup_{i\in A}\sim_i$, $\sim_A^D=\bigcap_{i\in A}\sim_i$, $\sim_A^C =(\sim_A^E)^+$ (note that $(\sim_A^E)^+$ is the transitive closure of $\sim_A^E$).
$\opK_i \varphi, \opE_A\varphi, \opD_A\varphi$ and $\opC_A\varphi$ respectively denote that $i$ knows, every agent in $A$ knows, agents in $A$ have common knowledge and agents in $A$ have distributed knowledge,
on the fact $\varphi$. The ATL (resp. ATL$^\ast$) logic extended with these epistemic modalities is called ATLK (resp. ATLK$^\ast$) logic. Given a state $s\in S$ and a binary relation $\backsimeq\subseteq S\times S$, we denote by $[s]^\backsimeq$ the set $\{s'\in S\mid s\backsimeq s' \}$.


  \begin{algorithm}[t]
   \caption{Function {\tt MC}$(\M,\varphi)$ outputs $\llbracket \varphi\rrbracket_\M$}
   \label{alg}
    \SetKwSwitch{Switch}{Case}{Other}{switch}{:}{case}{otherwise}{endcase}

    \Switch{$\varphi$}{
      \lCase{$q$}{\Return{$\{s\in S\mid q\in\lambda(s)\}$}}
      \lCase{$\neg \varphi'$}{\Return{$S\setminus {\tt MC}(\M,\varphi')$}}
      \lCase{$\varphi_1\wedge\varphi_2$}{\Return{${\tt MC}(\M,\varphi_1)\cap {\tt MC}(\M,\varphi_2)$}}
      \lCase{$\opK_i \varphi'$}{\Return {$\{s\in S \mid [s]^{\sim_i}\subseteq {\tt MC}(\M,\varphi')\}$}}
      \lCase{$\opE_A \varphi'$}{\Return {$\{s\in S \mid [s]^{\sim_A^E}\subseteq {\tt MC}(\M,\varphi')\}$}}
      \lCase{$\opD_A \varphi'$}{\Return {$\{s\in S \mid [s]^{\sim_A^D}\subseteq {\tt MC}(\M,\varphi')\}$}}
      \lCase{$\opC_A \varphi'$}{\Return {$\{s\in S \mid [s]^{\sim_A^C}\subseteq {\tt MC}(\M,\varphi')\}$}}
      \Case{$\opA{A}\phi$}{
        \ForEach{sub-state-formula $\varphi'$ \emph{\textbf{in}} $\phi$}{
          Replace $\varphi'$ by a fresh atomic proposition $q_{\varphi'}$ in $\varphi$,
         and let $\lambda(q_{\varphi'}):={\tt MC}(\M,\varphi')$\;
        }
        Compute $\llbracket \opA{A}\phi\rrbracket_\M$ by Lemma~\ref{lem:guessIr} or~\ref{lem:sat2pg}\;
        \Return{$\llbracket \opA{A}\phi\rrbracket_\M$}\;
      }
    }
 \end{algorithm}

By Lemma~\ref{lem:guessIr} and Lemma \ref{lem:sat2pg}, the model-checking problem for ATLK and ATLK$^\ast$ on \STCGS can be solved in PSPACE and 2EXPTIME respectively. As the model-checking problem of ATL$^\ast$ on CGS under $\IR$ setting is 2EXPTIME-complete~\cite{AHK02}, we have that

 \begin{theorem}
   The model-checking problem for ATLK$^\ast$ (resp. ATLK) on \STCGS is 2EXPTIME-complete (resp. in PSPACE).
 \end{theorem}

%

\section{Implementation and Experiments} \label{sec:imple}
We have implemented the ATLK/ATLK$^\ast$ model-checking algorithms in MCMAS~\cite{LomuscioQR17}. 
We conducted experiments on the castle game (CG)~\cite{PBJ14}.  
All experiments were conducted on a computer with 1.70GHz Intel Core E5-2603 CPU and 32GB of memory.


 In this CG game, there are several agents modelling workers and an environment agent. Each worker works for the benefit of a castle, and the environment keeps track of the Health Points (HP) of castles. Each castle preserves an HP valued up to 3, and 0 means it's defeated. Workers are able to attack a castle which they don't work for, or defend the castle which they work for, or do nothing. The castle gets damaged if the number of attackers is greater than the number of defenders, and the differences influence its HP.
 In this model, the number of states is $8000\times 4n$, the environment agent has 1 local action,
 and each worker agent has 4 local actions, where $n$ denotes the number of workers.

In this experiment, we consider \STCGS consisting of three worker agents $w_1,w_2,w_3$ and an environment agent $e$, where worker $w_i$ works for the castle $c_i$.
 \begin{itemize}
   \item $\varphi_1\equiv\opA{\{w_1,w_2\}}\opF (castle3Defeated)$: expresses that workers $w_1$ and $w_2$ can make castle $c_3$ defeated, no matter which strategies other agents use.
   \item $\varphi_2\equiv\opA{\{w_1,w_2\}}\opF (allDefeated)$: expresses that workers $w_1$ and $w_2$ can make all the castles defeated, no matter which strategies other agents use.
 \end{itemize}

 \begin{table}[t]
 \caption{Results of the castle game}
 \label{tab:CG}
 \setlength{\tabcolsep}{1pt}	
  \begin{tabular}{c|ccc|ccc}
   \hline
    \multirow{2}{*}{$\pi$} & \multicolumn{3}{c|}{$\varphi_1$} & \multicolumn{3}{c}{$\varphi_2$} \\
    \cline{2-7} &Lem.~\ref{lem:guessIr}  & Lem.~\ref{lem:sat2pg} & SAT & Lem.~\ref{lem:guessIr}  & Lem.~\ref{lem:sat2pg} & SAT \\ \hline
    $(\IR,\IR,\IR,\IR)$ & N/A & 20.295 &Y & N/A & 18.178 &Y\\
   $(\IR,\IR,\IR,\ir)$ & N/A & 7523.67 &Y& N/A & 7377.44 &Y\\
   $(\IR,\IR,\ir,\IR)$ & N/A & 31.904 &Y& N/A & 30.578 &N\\
   $(\IR,\ir,\IR,\IR)$ & N/A & 32.446 &Y& N/A & 31.259 &N\\
   $(\IR,\IR,\ir,\ir)$ &N/A & 3402.56 &Y& N/A & 3451.59&N \\
   $(\IR,\ir,\IR,\ir)$ & N/A & 3294.51&Y & N/A & 3366.71&N \\
   $(\IR,\ir,\ir,\IR)$ & 5.822 & 24.254&Y & 77.514 & 23.37&N \\
   $(\IR,\ir,\ir,\ir)$ & 13.791 & 113.493&Y & 45.679 & 113.647 &N\\ \hline

\end{tabular}
 \end{table}

The results are shown in Table~\ref{tab:CG}, where $(\sigma_1,\sigma_2,\sigma_3,\sigma_4)$ in each row denotes the strategy types of agents $e,w_1,w_2,w_3$, N/A denotes timeout (2.5 hours), Y (resp. N) denotes that the model satisfies (resp. fails) the formula, and columns 2--4 (resp. 5--7) show total time (in seconds) and result of verifying $\varphi_1$ (resp. $\varphi_2$) using Lemma~\ref{lem:guessIr} and Lemma~\ref{lem:sat2pg} respectively.

We observe that: (1) the strategy types of agents do affect the performance and results. In particular, the time significantly increases when $w_3$ is $\ir$-typed while $w_1$ or $w_2$ is $\IR$-typed; (2) Lemma~\ref{lem:guessIr} is more efficient when both $w_1$ and $w_2$ are $\ir$-typed; otherwise Lemma~\ref{lem:sat2pg} is more efficient.
This is because the number of possible strategies of $w_1$ and $w_2$ is small (using Lemma~\ref{lem:guessIr}) if
both $w_1$ and $w_2$ are $\ir$-typed.

\textbf{Further experiments (on the Book Store Application and Dining Cryptographers Protocol~\cite{LomuscioQR17}) can be found in~\cite{ACGStool19}.}

\section{Related Work}
The family of alternating-time temporal logics ATL, ATL$^\ast$ and alternating $\mu$-calculus (AMC) \cite{AHK02} for reasoning about games was introduced with motivations partially from MAS. Model-checking algorithms were also given with $\IR$-strategies.
\cite{HW03} extended ATL with knowledge operators and proposed corresponding model-checking algorithms. In their work, epistemic accessibility relations are considered in the interpretation of knowledge operators, but not for the strategies and outcomes. This means that agents still use $\IR$ strategies for coalition modalities $\opA{A}\psi$. This issue was discussed in~\cite{Jam03} which proposed an idea of $\iR$-strategies.
\cite{Sch04} introduced the notion of imperfect recall into ATL/ATL$^\ast$, and investigated
the model-checking problem for ATL/ATL$^\ast$ under four different strategic types.
Importantly, with $\iR$-strategies the model-checking problem becomes undecidable~\cite{DT11}.
%
\cite{BJ11} studied the semantics of AMC and proposed a model-checking algorithm
for the alternation-free fragment under the imperfect information setting. 
\cite{BJ14} further conducted a comprehensive comparison of variants of ATL/ATL$^\ast$ with different strategic types. The study corroborate that the agents' abilities play a prominent role in logic semantics.


In the previous work, strategies are revocable, i.e., when it comes to achieve
a goal in the (nested) sub-formula, previously selected strategies are deleted.
\cite{AGJ07} introduced a variant of ATL with \emph{irrevocable} strategies under the imperfect recall setting.
It was generalized into ATL/ATL$^\ast$ with strategy contexts~\cite{LM15}, which allows
agents to drop or inherit previously selected strategies.

Two strategic logics were introduced by~\cite{CHP10} and~\cite{MMPV14} and the model-checking problem was investigated therein under the $\IR$-setting.
Strategic logics extend LTL with first-order quantifications over strategies which naturally captures the multi-player game nature in the evolution of MAS. \cite{CermakLMM14} introduced knowledge operators in the strategic logic of~\cite{MMPV14} and proposed a model-checking algorithm with $\ir$-strategies. Here all agents must take $\ir$-strategies (so the potential inconsistency can be ruled out), but no other abilities are considered.
To gain decidability under $\iR$-setting, specific restrictions on the abilities of the agents were proposed ~\cite{BMM17,BMMRV17,BLMR17,BLMR17b,BLMR18}.



Our work is orthogonal to the existing work which defines the agents' abilities at the semantics level, but takes a more syntactic level by strengthening the model.


\section{Conclusion and Future Work} \label{sec:conc}
In this paper, we discussed the problem of existing semantics of ATL/ATL$^\ast$, and advocated the approach to make agents' abilities explicit in modeling. For this purpose, we introduced an extension of standard \CGS model, i.e. \STCGS, which define agents' abilities at the syntactic level of the system model. We explored the effects of strategy types in the semantics, in particular model-checking, of ATL/ ATL$^\ast$ over \STCGS, 
and provided model-checking algorithms with identified complexity. The algorithms are implemented in a tool MCMAS-ACGS, which has been applied to several applications to demonstrate the feasibility of the approach. 
This work represents the first systematic study towards different agents' abilities at the syntactic level, which is in contrast to the previous approaches at the semantic level. 

Currently we use ATL/ATL$^\ast$ as the specification, but the methodology can be extended to other logics such as Strategy Logic, and other agents' abilities such as strategy contexts. Several questions are left open such as axiomatization and satisfiability problem. We leave them for future work.


 \newpage
 \clearpage

 \bibliographystyle{named}
 \bibliography{references}

\begin{thebibliography}{}

\bibitem[\protect\citeauthoryear{{\AA}gotnes \bgroup \em et al.\egroup
  }{2007}]{AGJ07}
Thomas {\AA}gotnes, Valentin Goranko, and Wojciech Jamroga.
\newblock Alternating-time temporal logics with irrevocable strategies.
\newblock In {\em Proceedings of the 11th Conference on Theoretical Aspects of
  Rationality and Knowledge {(TARK)}}, pages 15--24, 2007.

\bibitem[\protect\citeauthoryear{Alur \bgroup \em et al.\egroup }{2002}]{AHK02}
R.~Alur, T.~A. Henzinger, and O.~Kupferman.
\newblock Alternating-time temporal logic.
\newblock {\em Journal of the {ACM}}, 49(5):672--713, 2002.

\bibitem[\protect\citeauthoryear{Belardinelli \bgroup \em et al.\egroup
  }{2017a}]{BLMR17}
Francesco Belardinelli, Alessio Lomuscio, Aniello Murano, and Sasha Rubin.
\newblock Verification of broadcasting multi-agent systems against an epistemic
  strategy logic.
\newblock In {\em Proceedings of the Twenty-Sixth International Joint
  Conference on Artificial Intelligence ({IJCAI})}, pages 91--97, 2017.

\bibitem[\protect\citeauthoryear{Belardinelli \bgroup \em et al.\egroup
  }{2017b}]{BLMR17b}
Francesco Belardinelli, Alessio Lomuscio, Aniello Murano, and Sasha Rubin.
\newblock Verification of multi-agent systems with imperfect information and
  public actions.
\newblock In {\em Proceedings of the 16th Conference on Autonomous Agents and
  MultiAgent Systems ({AAMAS})}, pages 1268--1276, 2017.

\bibitem[\protect\citeauthoryear{Belardinelli \bgroup \em et al.\egroup
  }{2018}]{BLMR18}
Francesco Belardinelli, Alessio Lomuscio, Aniello Murano, and Sasha Rubin.
\newblock Decidable verification of multi-agent systems with bounded private
  actions.
\newblock In {\em Proceedings of the 17th International Conference on
  Autonomous Agents and MultiAgent Systems ({AAMAS})}, pages 1865--1867, 2018.

\bibitem[\protect\citeauthoryear{Berthon \bgroup \em et al.\egroup
  }{2017a}]{BMM17}
Rapha{\"{e}}l Berthon, Bastien Maubert, and Aniello Murano.
\newblock Decidability results for {ATL}* with imperfect information and
  perfect recall.
\newblock In {\em Proceedings of the 16th Conference on Autonomous Agents and
  MultiAgent Systems ({AAMAS})}, pages 1250--1258, 2017.

\bibitem[\protect\citeauthoryear{Berthon \bgroup \em et al.\egroup
  }{2017b}]{BMMRV17}
Rapha{\"{e}}l Berthon, Bastien Maubert, Aniello Murano, Sasha Rubin, and
  Moshe~Y. Vardi.
\newblock Strategy logic with imperfect information.
\newblock In {\em Proceedings of the 32nd Annual {ACM/IEEE} Symposium on Logic
  in Computer Science ({LICS})}, pages 1--12, 2017.

\bibitem[\protect\citeauthoryear{Bulling and Jamroga}{2011}]{BJ11}
Nils Bulling and Wojciech Jamroga.
\newblock Alternating epistemic mu-calculus.
\newblock In {\em Proceedings of the 22rd International Joint Conference on
  Artificial Intelligence}, pages 109--114, 2011.

\bibitem[\protect\citeauthoryear{Bulling and Jamroga}{2014}]{BJ14}
Nils Bulling and Wojciech Jamroga.
\newblock Comparing variants of strategic ability: how uncertainty and memory
  influence general properties of games.
\newblock {\em Autonomous Agents and Multi-Agent Systems}, 28(3):474--518,
  2014.

\bibitem[\protect\citeauthoryear{Cerm{\'{a}}k \bgroup \em et al.\egroup
  }{2014}]{CermakLMM14}
Petr Cerm{\'{a}}k, Alessio Lomuscio, Fabio Mogavero, and Aniello Murano.
\newblock {MCMAS-SLK:} {A} model checker for the verification of strategy logic
  specifications.
\newblock In {\em Proceedings of the 26th International Conference on Computer
  Aided Verification ({CAV})}, pages 525--532, 2014.

\bibitem[\protect\citeauthoryear{Chatterjee \bgroup \em et al.\egroup
  }{2010}]{CHP10}
K.~Chatterjee, T.~A. Henzinger, and N.~Piterman.
\newblock Strategy logic.
\newblock {\em Information and Computation}, 208(6):677--693, 2010.

\bibitem[\protect\citeauthoryear{Clarke \bgroup \em et al.\egroup
  }{1983}]{CES83}
Edmund~M. Clarke, E.~Allen Emerson, and A.~Prasad Sistla.
\newblock Automatic verification of finite state concurrent systems using
  temporal logic specifications: {A} practical approach.
\newblock In {\em POPL}, pages 117--126, 1983.

\bibitem[\protect\citeauthoryear{Dima and Tiplea}{2011}]{DT11}
Catalin Dima and Ferucio~Laurentiu Tiplea.
\newblock Model-checking {ATL} under imperfect information and perfect recall
  semantics is undecidable.
\newblock {\em CoRR}, abs/1102.4225, 2011.

\bibitem[\protect\citeauthoryear{Jamroga and van~der Hoek}{2004}]{JH04}
Wojciech Jamroga and Wiebe van~der Hoek.
\newblock Agents that know how to play.
\newblock {\em Fundam. Inform.}, 63(2-3):185--219, 2004.

\bibitem[\protect\citeauthoryear{Jamroga}{2003}]{Jam03}
Wojciech Jamroga.
\newblock Some remarks on alternating temporal epistemic logic.
\newblock {\em Proceedings of Formal Approaches to Multi-Agent Systems
  ({FAMAS})}, pages 133--140, 2003.

\bibitem[\protect\citeauthoryear{Jurdzinski}{2000}]{Jur00}
Marcin Jurdzinski.
\newblock Small progress measures for solving parity games.
\newblock In {\em Proceedings of the 17th Symposium on Theoretical Aspects of
  Computer Science}, pages 290--301, 2000.

\bibitem[\protect\citeauthoryear{Laroussinie and Markey}{2015}]{LM15}
Fran{\c{c}}ois Laroussinie and Nicolas Markey.
\newblock Augmenting {ATL} with strategy contexts.
\newblock {\em Inf. Comput.}, 245:98--123, 2015.

\bibitem[\protect\citeauthoryear{Laroussinie \bgroup \em et al.\egroup
  }{2008}]{LMO08}
Fran{\c{c}}ois Laroussinie, Nicolas Markey, and Ghassan Oreiby.
\newblock On the expressiveness and complexity of {ATL}.
\newblock {\em Logical Methods in Computer Science}, 4(2), 2008.

\bibitem[\protect\citeauthoryear{Lomuscio \bgroup \em et al.\egroup
  }{2017}]{LomuscioQR17}
Alessio Lomuscio, Hongyang Qu, and Franco Raimondi.
\newblock {MCMAS:} an open-source model checker for the verification of
  multi-agent systems.
\newblock {\em {STTT}}, 19(1):9--30, 2017.

\bibitem[\protect\citeauthoryear{MCMAS-ACGS}{2018}]{ACGStool19}
MCMAS-ACGS.
\newblock Open source tool.
\newblock \url{https://github.com/MCMAS-ACGS}, 2018.

\bibitem[\protect\citeauthoryear{Mogavero \bgroup \em et al.\egroup
  }{2014}]{MMPV14}
Fabio Mogavero, Aniello Murano, Giuseppe Perelli, and Moshe~Y. Vardi.
\newblock Reasoning about strategies: On the model-checking problem.
\newblock {\em {ACM} Transations on Computational Logic}, 15(4):34:1--34:47,
  2014.

\bibitem[\protect\citeauthoryear{Pilecki \bgroup \em et al.\egroup
  }{2014}]{PBJ14}
Jerzy Pilecki, Marek~A. Bednarczyk, and Wojciech Jamroga.
\newblock Model checking properties of multi-agent systems with imperfect
  information and imperfect recall.
\newblock In {\em Proceedings of the 7th International Conference on
  Intelligent Systems ({IS})}, pages 415--426, 2014.

\bibitem[\protect\citeauthoryear{Piterman}{2006}]{Pit06}
N.~Piterman.
\newblock From nondeterministic {B{\"u}chi} and {Streett} automata to
  deterministic parity automata.
\newblock In {\em LICS 2006}, pages 255--264, 2006.

\bibitem[\protect\citeauthoryear{Schobbens}{2004}]{Sch04}
Pierre{-}Yves Schobbens.
\newblock Alternating-time logic with imperfect recall.
\newblock {\em Electr. Notes Theor. Comput. Sci.}, 85(2):82--93, 2004.

\bibitem[\protect\citeauthoryear{van~der Hoek and Wooldridge}{2003}]{HW03}
Wiebe van~der Hoek and Michael Wooldridge.
\newblock Cooperation, knowledge, and time: Alternating-time temporal epistemic
  logic and its applications.
\newblock {\em Studia Logica}, 75(1):125--157, 2003.

\end{thebibliography}


 \newpage
 \clearpage

 \appendix
 \onecolumn

\section{Proof of Theorem~\ref{thm:und}}
Given a Turing machine $M$, let $\G=(S, \{s_0\}, Ag, (Ac_i)_{i\in Ag}, (\sim_i)_{i\in Ag}, (P_i)_{i\in Ag}, \Delta, \lambda )$ be the CGS constructed as in~\cite{DT11} and $s_0$ be an initial state in $\G$ such that $\G, s_0 \models_{\iR} \opA{\{1,2\}}\opG \ ok$ iff $M$ does not halt on the empty word. Let $\M=(\G,\pi)$ be an \STCGS such that for every agent $i$ in $\M$, $\pi(i)=\iR$ if $i\in\{1,2\}$, otherwise $\pi(i)=\IR$. Therefore,  $\G, s_0 \models_{\iR} \opA{\{1,2\}}\opG \ ok$ iff $\M, s_0 \models \opA{\{1,2\}}\opG \ ok$. The result immediately follows.

\section{Proof of Proposition~\ref{prop:Isemid}}
Without loss of generality, we assume that for every $i\in Ag$, $\pi(i)\neq\Ir$, as $\Ir$ can be seen as a special case of $\ir$. We first construct the tree-unfolding of $\M$ from the state $s$.

 The tree-unfolding of $\M$ from $s$ is an \STCGS $\M^*_s= (\G^*, \pi^*)$ such that $\G^*=(S^+, S_0, Ag, (Ac_i)_{i\in Ag}, (\sim_i^*)_{i\in Ag}, (P_i^*)_{i\in Ag}, \Delta^*, \lambda^* )$, where
 \begin{itemize}
   \item for every $i\in Ag$, $\pi^*(i)=\Ir$ if $\pi(i)=\IR$ and $i\in A$, otherwise $\pi^*(i)=\pi(i)$;
   \item for every $i\in Ag$ and $\rho_1,\rho_2\in S^+$, $\rho_1 \sim_i^* \rho_2$, if either $\last(\rho_1)\sim_i \last(\rho_2)$ and $\pi(i)\neq \IR$, or $\rho_1= \rho_2$ and $\pi(i)= \IR$,
   \item $\lambda^*(\rho)=\lambda(\last(\rho))$ for every $\rho\in S^+$;
   \item $P_i^*(\rho)=P_i(\last(\rho))$ for every $i\in Ag$ and $\rho\in S^+$;
   \item $\Delta^*(\rho,\vec{a})=\rho \cdot \Delta(\last(\rho),\vec{a})$ for every $\rho\in S^+,\vec{a}\in Ac$.
 \end{itemize}

 From the above definition, we can immediately get that the tree-unfolding $\M^*_q$ is a tree-like \STCGS, namely, every state can be reached by a unique finite path from the root. $\IR$-strategies of $A$ from $s$ in $\M$ correspond exactly to $\Ir$-strategies of $A$ in the tree unfolding $\M^*_q$ from $s$, while the types of other agents are same under $\pi$ and $\pi^*$. Thus, $\M, s \models \opA{A}\varphi$ iff  $\M^*_s, s \models \opA{A}\varphi$. Note that this result does not hold if $\varphi$ is a general LTL formula.

 Next, we will show that $\M', s \models \opA{A}\varphi$ iff  $\M^*_s, s \models \opA{A}\varphi$.

 $(\Rightarrow)$ Suppose $\M', s \models \opA{A}\varphi$, let $\xi_A$ be the collective strategy such that for every path $\rho\in \outcomes_{\M'}(s,\xi_A)$, $\M',\rho\models \varphi$. Let $\xi_A^*$ be the function such that for every $i\in A$ and $\rho\in S^+$, $\xi_A^*(i)(\rho)=\xi_A(i)(\last(\rho))$.

 First, we show that $\xi_A^*$ is a collective strategy of $A$ in $\M^*_s$. Consider an agent $i\in A$ and two states $\rho_1,\rho_2\in S^+$ such that $\rho_1\sim_i^*\rho_2$, if $\pi(i)\neq \IR$, then $\last(\rho_1)\sim_i \last(\rho_2)$ which implies that $\xi_A(i)(\last(\rho_1))=\xi_A(i)(\last(\rho_2))$, hence $\xi_A^*(i)(\rho_1)=\xi_A^*(i)(\rho_2)$. Otherwise $\rho_1= \rho_2$ and $\pi(i)= \IR$. This implies that $\xi_A(i)(\last(\rho_1))=\xi_A(i)(\last(\rho_2))$, hence $\xi_A^*(i)(\rho_1)=\xi_A^*(i)(\rho_2)$ as well. Therefore, $\xi_A^*$ is a collective strategy of $A$ in $\M^*_s$.

 Next, we show that for every collective strategy $\xi_{\nA}^*$ of $\nA$ in $\M^*_s$, $\play(s,\xi_A^*,\xi_{\nA}^*)\models \varphi$. Suppose $\play(s,\xi_A^*,\xi_{\nA}^*)=\rho_0\rho_1\cdots$. Let $\xi_{\nA}$ be the function such that for every $i\in \nA$ and $j\geq 0$, $\xi_A(i)(\last(\rho_j))=\xi_A^*(i)(\rho_j)$ if $\pi(i)\neq\IR$, otherwise $\xi_A(i)(\rho_j)=\xi_A^*(i)(\rho_0\cdots\rho_j)$. Consider $j,k\geq 0$ such that $\last(\rho_j)\sim_i\last(\rho_k)$ for some $i\in \nA$, if $\pi(i)\neq \IR$, then $\rho_j\sim_i^*\rho_k$, which implies that $\xi_A^*(i)(\rho_j)=\xi_A^*(i)(\rho_k)$, hence $\xi_A(i)(\last(\rho_j))=\xi_A(i)(\last(\rho_k))$. Otherwise, $\pi(i)=\IR$, $i$ can choose any action at any state of $\rho_j$. Thus, $\xi_{\nA}$ is a collective strategy of $\nA$ in $\M'$ and $\play(s,\xi_A,\xi_{\nA})=\last(\rho_0)\last(\rho_1)\cdots$. The result immediately follows from the fact that $\lambda^*(\rho)=\lambda(\last(\rho))$ for every $\rho\in S^+$.

 $(\Leftarrow)$ Suppose $\M^*_s, s \models \opA{A}\varphi$, let $\xi_A^*$ be the collective strategy such that for every path $\rho\in \outcomes_{\M^*_s}(s,\xi_A^*)$, $\M^*_s,\rho\models \varphi$. We assume that there is a total order $\preceq$ on set $S^+$, and denote by $\min(U)$ the minimal one of the set of states $U\subseteq S^+$ with respect to the order $\preceq$. Let $\xi_A$ be the function such that for every $i\in A$ and $s'\in S$, $\xi_A(i)(s')=\xi_A^*(i)(\min(\{\rho\in S^+\mid \last(\rho)=s'\}))$.

 First, we show that $\xi_A$ is a collective strategy of $A$ in $\M'$. Consider an agent $i\in A$ and two states $s_1,s_2\in S$ such that $s_1\sim_i s_2$, if $\pi(i)\neq \IR$, then for each pair of states $\rho_1,\rho_2\in S^+$ such that $\last(\rho_1)=s_1$ and $\last(\rho_2)=s_2$, we have $\rho_1\sim_i^* \rho_2$, which implies that $\xi_A^*(i)(\rho_1)=\xi_A^*(i)(\rho_2)$, hence $\xi_A(i)(s_1)=\xi_A(i)(s_2)$. Otherwise $s_1= s_2$ and $\pi(i)= \IR$. We choose $\xi_A(i)(s_1)=\xi_A(i)(s_2)=\xi_A^*(i)(\min(\{\rho\in S^+\mid \last(\rho)=s_1\}))$. Therefore, $\xi_A$ is a collective strategy of $A$ in $\M'$.

 Consider a collective strategy $\xi_{\nA}$ of $\nA$ in $\M'$, let $\rho=\play(s,\xi_A,\xi_{\nA})$, then we have $\rho_{[0..0]}\rho_{[0..1]}\rho_{[0..2]}\cdots \in\outcomes_{\M^*_s}(s,\xi_A^*)$. The result immediately follows from the fact that $\lambda^*(\rho)=\lambda(\last(\rho))$ for every $\rho\in S^+$.
 \qed

Recalling that~\cite{AHK02} observed that both semantics of ATL under $\Ir$-strategies and $\IR$-strategies coincide for $\CGS$.
This result was generalized and formally proved in infinite $\CGS$ (i.e., no finiteness with respect to the set of states and actions) (cf. Proposition 1~\cite{BJ14}).
Proposition~\ref{prop:Isemid} can be seen as a generalization of the result of~\cite{AHK02} and
could be extended to the infinite \STCGS similar to~\cite{BJ14}.

\section{Effects of Strategy Types}
 \label{sec:stcgsvscgs}

 By restricting all the strategy types to $\IR$, straightforwardly we have:

 \begin{proposition}\label{prop:IRsem}
   Let $\M=(\G,\pi)$ be an \STCGS where for each $i\in Ag$, $\pi(i)=\IR$.  For each state $s$ of $\M$ and ATL$^\ast$ formula $\varphi$,
   $\G, s \models_{\IR} \varphi$ iff $\M, s \models \varphi$.
 \end{proposition}
\emph{Proof}. By applying structural induction, it suffices to show that the result holds for formulae of the form $\opA{A}\phi$. By applying induction hypothesis, for every path $\rho$, the following holds: $\G, \rho\models_{\IR} \varphi$ iff $\M, \rho \models \varphi$. For each pair $(\xi_A,\upsilon_A^\sigma)$ of collective strategies such that $\xi_A=\upsilon_A^\sigma$, $\outcomes_\M(s,\xi_A) = \outcomes_\G^{\sigma}(s, \upsilon_A^\sigma)$. Each $i\in A$ has same sets of possible $\IR$-strategies in $\G$ and $\M$, hence $\G, \rho\models_{\IR} \opA{A}\phi$ iff $\M, \rho \models \opA{A}\phi$.
 \qed

In light of Proposition~\ref{prop:stcgs2cgs} and Proposition~\ref{prop:IRsem}, we shall investigate the effects of strategy types by considering \STCGS with various different setups of strategy types.

Given a set $A$ of agents, for two functions $\pi_1,\pi_2:Ag\rightarrow \StrT$, $\pi_1$ is
\emph{coarser} than $\pi_2$ with respect to $A$, denoted by $\pi_1\preceq_A \pi_2$, if for every $i\in A$, $\pi_1(i)=\pi_2(i)$ and for every $j\in \nA$, one of the following conditions holds:

 \begin{itemize}
   \item  $\pi_1(j)=\IR$,  $\pi_2(j)=\IR$;
   \item  $\pi_1(j)=\Ir$,  $\pi_2(j)\in\{\IR,\Ir\}$;
   \item  $\pi_1(j)=\iR$,  $\pi_2(j)\in\{\IR,\iR\}$;
   \item  $\pi_1(j)=\ir$,  $\pi_2(j)\in\{\IR,\Ir,\iR,\ir\}=\StrT$.
 \end{itemize}

 \begin{lemma}\label{lem:out2out}
   Let $A$ be a set of agents and $s$ be a state of a \CGS $\G$. For two functions $\pi_1,\pi_2:Ag\rightarrow \StrT$ with $\pi_1\preceq_A \pi_2$, and any collective strategy $\xi_A$ of $A$, we have:
   \[
      \outcomes_{(\G,\pi_1)}(s,\xi_A) \subseteq \outcomes_{(\G,\pi_2)}(s,\xi_A).
   \]
 \end{lemma}

 Lemma~\ref{lem:out2out} reveals the effect of strategy types of $\nA$ on the outcomes.
 It is easy to observe that if $\pi_2(i)=\sigma$ for all $i\in A$, then for every collective $\sigma$-strategy $\upsilon_A^\sigma$ such that $\xi_A=\upsilon_A^\sigma$, we have that $\outcomes_{(\G,\pi_2)}(s,\xi_A)\subseteq \outcomes_\G^{\sigma}(s, \upsilon_A^\sigma)$. Moreover, if $\pi_2(i)=\IR$ for all $i\in\nA$, then $\outcomes_{(\G,\pi_2)}(s,\xi_A)= \outcomes_\G^{\sigma}(s, \upsilon_A^\sigma)$.

 An ATL$^\ast$ formula $\varphi$ is   \emph{positive} if (1) for each occurrence of $\opA{A}\phi$ in $\varphi$, $\phi$ is an LTL formula, (2) there is no occurrence of $\opUA{A}\phi$ in $\varphi$, and (3) negations $\neg$ only appear in front of atomic propositions.
 For example, $\opA{A}\opX \ q$ is positive, while $\neg\opA{A}\opX \ q$ is not  positive.
 Given a formula $\varphi$, let $Ag_\varphi$ denote the set of agents that appear in $\varphi$.
 By Lemma~\ref{lem:out2out}, we have:

 \begin{proposition}\label{prop:sem}
   Let $\G$ be a \CGS, $s$ be a state of $\G$ and $\varphi$ be a positive ATL/ATL$^\ast$ formula. For  $\pi_1,\pi_2:Ag\rightarrow \StrT$ such that $\pi_1\preceq_{Ag_\varphi} \pi_2$,
if $(\G,\pi_2), s \models \varphi$, then $(\G,\pi_1), s \models \varphi$.

 \end{proposition}
\emph{Proof}.
By applying structural induction, it suffices to show that the result holds for formulae of the form $\opA{A}\phi$. We suppose $(\G,\pi_2), s \models \opA{A}\phi$, otherwise Item 1 immediately holds.

 Since $(\G,\pi_2), s \models \opA{A}\phi$, then there exists a collective strategy  $\xi_A$ of agents $A$ such that for each path $\rho\in \outcomes_{(\G,\pi_2)}(s,\xi_A)$, $(\G,\pi_2), \rho \models \phi$ holds. Since $A\subseteq Ag_\varphi$ and for every $i\in Ag_\varphi$, $\pi_1(i)=\pi_2(i)$ and $\pi_1\preceq_{Ag_\varphi} \pi_2$, then $\pi_1\preceq_{A} \pi_2$. By Lemma~\ref{lem:out2out}, we get that $\outcomes_{(\G,\pi_1)}(s,\xi_A) \subseteq \outcomes_{(\G,\pi_2)}(s, \xi_A)$.

 By applying induction hypothesis, for every state formula $\psi$ in $\phi$ and state $s'$ of $\G$, if $(\G,\pi_2), s' \models \psi$, then $(\G,\pi_1), s' \models \psi$. Therefore, for each path $\rho\in\outcomes_{(\G,\pi_1)}(s,\xi_A)$, we have $(\G,\pi_1), \rho \models \phi$. The result follows.

 \qed

For positive ATL/ATL$^\ast$ formulae $\varphi$, even if the agents of $Ag_\varphi$ have the same strategy types in \STCGS $(\G,\pi)$ and \CGS $\G$,  verifying $\G$ against $\varphi$ under $\sigma$ will examine more behavior than verifying $(\G,\pi)$ against $\varphi$, where $i\in Ag_\varphi$ and $\pi(i)=\sigma$.
Therefore, if the behavior of a MAS is exactly modeled as an \STCGS $\M$ rather than a \CGS $\G$ with strategy type $\sigma$,
verifying  $\G$ against $\varphi$ under $\sigma$ may lead to incorrect result. However, more restrictions on strategy types and ATL/ATL$^\ast$ formulae can make them coincide, as the following proposition shows.

 \begin{proposition}\label{prop:IRsemid}
   Let $s$ be a state of $\M=(\G,\pi)$ and $\sigma\in\StrT$ be a strategy type. Assume ATL/ATL$^\ast$ formula $\varphi$ satisfies
   (1) for every $i\in Ag_\varphi$, $\pi(i)=\sigma$,
   (2) for every $i\in Ag\setminus Ag_\varphi$, $\pi(i)=\IR$,
   and (3) for every occurrence of $\opA{A'}\phi$ in $\varphi$, $Ag_\varphi=A'$. Then we have $\G, s \models_\sigma \varphi$ iff $\M, s \models \varphi$.
 \end{proposition}

\section{More Experimental Results}
\subsection{Experiment on Dining Cryptographers Protocol}
Dining cryptographers protocol is one of anonymity protocols aimed at establishing
the privacy of principals during an exchange~\cite{LomuscioQR17}.
The dining cryptographers protocol can be modeled as MAS.
In this game, $n$ cryptographers share a meal around a circular table.
Either one of them or their employer paid for the meal.
They are curious whether it was sponsored by their employer without revealing the
identity of the payer (if one of them did pay).
The protocol works as follows: each cryptographer 1) tosses a coin and shows the outcome
to his right-hand neighbour, 2) announces whether the two coins agree or not if he/she
is not payer, otherwise announces
the opposite of what he/she sees. Their employer
is the payer if an even number of cryptographers claiming
that the two coins are different, otherwise not.
For experimental purpose, we allow the cryptographer who paid for the meal
announces either the two coins agree or not no matter what he/she saw.

  \begin{table}[t]
  \caption{Results of dining cryptographers protocol}
  \label{tab:dining}
  \centering
  \begin{tabular}{cc|ccc|ccc}
   \hline
   \multirow{2}{*}{\#Crypts} &  \multirow{2}{*}{\#States} & \multicolumn{3}{c|}{Lemma~\ref{lem:guessIr}} & \multicolumn{3}{c}{Lemma~\ref{lem:sat2pg}}  \\    \cline{3-8}
                            &                            & $\psi_1$ & $\psi_2$ & $\psi_3$ &   $\psi_1$ & $\psi_2$ & $\psi_3$  \\  \hline
   3& 160 & 0.022 & 0.016 & 0.013 & 6.439 & 5.838 & 5.852 \\
   4& 384 & 0.059 & 0.049 & 0.028 & 6.928 & 6.744 & 7.242 \\
   5& 896 & 0.133 & 0.114 & 0.049 & 8.839 & 8.874 & 8.88 \\
   6& 2048 & 0.315 & 0.328 & 0.163 & 12.567 & 12.724 & 12.865 \\
   7& 4608 & 0.929 & 1.388 & 0.382 & 22.938 & 23.411 & 23.654 \\
   8& 10240 & 3.463 & 4.022 & 0.834  & 60.642 & 60.583 & 63.064 \\
   9& 22528 & 9.19 & 8.913 & 1.721  & 266.844 & 240.003 & 254.293 \\
   10& 49152 & 21.988 & 21.927 & 5.094 & 1712.62 & 1588.06 & 1762.88 \\    \hline
  \end{tabular}
 \end{table}


In this experiment, $n$ ranges from $3$ to $10$,
a cryptographer whose is payer and one of other cryptographers use $\ir$-strategies, the others all use $\IR$-strategies.
We verify three formulae $\psi_1$, $\psi_2$ and $\psi_3$, where  $\psi_i$ expresses that if the number of ``saydifferent" is odd and the $i$-th cryptographer is not the payer,
then he/she knows that the bill is paid by one of the others, but cannot tell exactly who is the payer.
For instance, in the three cryptographers case,
\[\psi_1\equiv\opA{\emptyset}\opG( (odd\wedge \neg c1paid)\rightarrow
       ((\opK_{c1}(c2paid\vee c3paid))
       \wedge \neg \opK_{c1} c2paid
       \wedge \neg \opK_{c1} c3paid)).\]

The results are shown in Table~\ref{tab:dining}, where column 1 gives the number of cryptographers,
column 2 gives the number of states, columns 3--5 (resp. columns 6--7) show the total time of respectively verifying $\psi_1$, $\psi_2$ and $\psi_3$ using Lemma~\ref{lem:guessIr} (resp. Lemma~\ref{lem:sat2pg}).
Both $\psi_1$ and $\psi_2$ are satisfied by all the models, while $\psi_3$ not.
We observe that Lemma~\ref{lem:guessIr} is more efficient than Lemma~\ref{lem:sat2pg}, as the coalitions in all the formulae are $\emptyset$.
From this experiment, one may conclude the reasonable scalability of our tool.

\subsection{Experiment on Book Store Scenario (BSS)}
 The BSS model depicts a deal between two agents: a supplier (S) and a purchaser (P)~\cite{LomuscioQR17}.
 Initially, S is waiting for an order from P, and P is ready for initiating a trade.
 Upon receiving an order of some e-good from P, S can make a decision whether accepts the order or not do, and later notify P.
 If S accepts, then P can pay fee. Once paid, S can either reject the payment or accept and deliver the good.
 If P received the good, then trade is completed.
During the trade, P can revoke the order, both S and P can terminate the trade, after which
the information of the trade should be symmetric at any time.
In this model, the S has 15  local states and 13 actions, and P has 12  local states and 7 actions.
In this experiment, we verify the model against the following formulae. 

\[\varphi_1\equiv\opA{\emptyset}\opG ((S\&P\_no\_T)\rightarrow (\opK_S \opA{\{S,P\}}\opF \ trade\_end))\]
expresses that ``if neither S nor P terminates the trade (i.e., $S\&P\_no\_T$ is ture), then S knows that they can cooperatively complete the trade eventually" always holds.

\[\varphi_2\equiv\opA{\{S,P\}}(S\&P\_no\_T  \ \opU (trade\_end\wedge \neg trade\_success))\] expresses that the trade can ends by P asking for refund.

 \begin{table}
  \caption{Results of BSS}
   \label{tab:bookstore}
   \centering
   \begin{tabular}{c|ccc|ccc}
    \hline
    \multirow{2}{*}{$\pi$} & \multicolumn{3}{c|}{$\varphi_1$} & \multicolumn{3}{c}{$\varphi_2$} \\ \cline{2-7}
     & Lemma~\ref{lem:guessIr}  & Lemma~\ref{lem:sat2pg} & SAT & Lemma~\ref{lem:guessIr}  & Lemma~\ref{lem:sat2pg} & SAT \\ \hline
    $(\IR,\IR)$ & 4.237 & 11.264 & Y & 0.08 & 5.566 & Y  \\
    $(\IR,\Ir)$ & 4.102 & 12.185 & Y & 0.081 & 5.124 & Y \\
    $(\IR,\ir)$ & 4.094 & 11.459 & Y & 0.081 & 5.26 & Y \\
    $(\Ir,\IR)$ & 4.095 & 17.398 & Y & 0.081 & 6.096 & Y \\
    $(\Ir,\Ir)$ & 4.086 & 30.649 & Y & 0.082 & 7.183 & Y \\
    $(\Ir,\ir)$ & 4.112 & 32.985 & Y & 0.082 & 8.009 & Y \\
    $(\ir,\IR)$ & 4.162 & 17.842 & N & 0.082 & 5.96  & Y \\
    $(\ir,\Ir)$ & 4.144 & 31.155 & N & 0.082 & 7.592 & Y \\
    $(\ir,\ir)$ & 4.157 & 30.73 & N & 0.082 & 7.473 & Y \\ \hline
  \end{tabular}
 \end{table}

The results are shown in Table~\ref{tab:bookstore}.  Each row presents the result of one of strategy type combinations of $S$ and $P$, for instance,
$(\IR,\ir)$ denotes that S has an $\IR$-strategy while P has $\ir$-strategy.
Columns 2-4 (resp. columns 5-7) show total time and result of verifying $\varphi_1$ (resp. $\varphi_2$) using Lemma~\ref{lem:guessIr} (resp. Lemma~\ref{lem:sat2pg}).
The results of $\varphi_1$ confirm that strategy types affect the truth of formulae.
Lemma~\ref{lem:guessIr} performs better than Lemma~\ref{lem:sat2pg} both on $\varphi_1$ and $\varphi_2$ in this experiment.

\end{document}